\documentclass[11pt,a4paper]{article}
\usepackage[hyperref]{acl2018}
\usepackage{times}
\usepackage{latexsym}
\usepackage{bbm}
\usepackage{url}
\usepackage{epsfig}
\usepackage{graphicx}
\usepackage{amsmath}
\usepackage{amssymb}
\usepackage{multirow}
\usepackage[
  separate-uncertainty = true,
  multi-part-units = repeat
]{siunitx}

\aclfinalcopy 

\title{Word Embedding Perturbation for Sentence Classification}

 \author{Dongxu Zhang, Zhichao Yang\\
         CS 682, UMass Amherst, Fall 2017\\     
        \url{{dongxuzhang,zhichaoyang}@cs.umass.edu}\\
        }

\date{}

\begin{document}
\maketitle

\begin{abstract}

In this technique report, we aim to mitigate the overfitting problem of natural language by applying data augmentation methods. Specifically, we attempt several types of noise to perturb the input word embedding, such as Gaussian noise, Bernoulli noise, and adversarial noise, etc.
We also apply several constraints on different types of noise.
By implementing these proposed data augmentation methods, 
the baseline models can gain improvements on several sentence classification tasks.

\end{abstract}

\section{Introduction}

Human annotation data is always insufficient in supervised learning problems. 
And the strong representative 
ability of neural networks usually makes the model easier to overfit on relatively small datasets. 
Data augmentation tackles this issue by 
automatically generating more training data with label-preserving transformations on the existing dataset. For example, 
~\newcite{krizhevsky2012imagenet} generated new images by randomly extracting slightly smaller
images from original ones or by adding minor changes on each color channel. 
This technique was also used in speech recognition~\cite{hannun2014deep,jaitly2013vocal} 
by changing the tones of acoustic signals or adding background noises.

In natural language domain, data augmentation is rarely utilized because words are discrete
and cannot be changed in a continuous space. 
There are some recent work in language domain which created more training data using additional domain knowledge.
~\newcite{zhang2015text} implemented word replacement using thesaurus.
In the relation extraction task, ~\newcite{xu2016improved} doubles the number of training
data by reversing the dependency
path between two entities. Though promising, these methods require extra 
knowledge bases or well-trained NLP tools.
~\newcite{iyyer2015deep} and ~\newcite{zhang2016learning}
applied random word dropout for bag-of-word models 
and accomplished improvements on text classification
tasks. 
However, they did not thoroughly apply and compare different noise strategies on the word embedding space.

In this paper, we focus on data augmentation for sentence 
classification tasks.
Instead of utilizing the discrete language space,
we will implement data modification on the continuous word 
embedding~\cite{mikolov2013distributed,pennington2014glove} space.
Specifically, we follow the conventional neural net setting of language processing~\cite{collobert2011natural} and
generate more training data by slightly modifying the
sequence of word embeddings. This word embedding sequence is also 
known as the word embedding layer which is produced
by looking up the word embedding matrix for each discrete input token. 

There are two possible strategies to change the word embedding sequence:
One is to add random noise without any domain knowledge; 
The other is to change the text using additional knowledge, 
such as extracting a sub-text or replacing an expression from the input sentence. 
Unlike images, language expressions are quite sensitive to individual words or clauses.  
In this case, for the second strategy, it requires high-performance models to accomplish the text modification 
process, or else it will generate incorrect biased training data with wrong labels.

Thus, in this paper, we choose the first strategy of adding small random noise on the word embedding layer, named \emph{word embedding perturbation}. 
As shown in the next section, the 
\emph{dropout}~\cite{srivastava2014dropout} method can be regarded as one type of noise and we will explore different noise types in this paper.
Briefly speaking, this paper has two main contributions:
\begin{itemize}
\item Present several perturbation methods on word embedding layer, such as Gaussian noise, 
Bernoulli noise and adversarial training. 
Particularly, adversarial training can improve sentence classification tasks constantly.
\item The performances can be boosted further by adding reasonable constraints over random noises, such as
spatial or loss-adversarial constraints.
\end{itemize}



\section{Technical Approach}

In this section, we will introduce different data augmentation approaches.
Specifically, we will first cover three basic types of noises such as Gaussian noise, Bernoulli Noise and Adversarial noise. 
Then we propose some variants of them by adding constraints among different word embedding features, 
such as word dropout, semantic dropout, Gaussian adversarial noise and Bernoulli adversarial noise.

\subsection{Gaussian Noise}
Gaussian distribution is one of the most natural choice for noise sampling. 
Here, we can simply add the noise matrix $e$ on the input word embedding sequence:

\begin{equation}
\label{eq:gaussian_mul}
X_{emb} \leftarrow X_{emb} \odot e, e \sim \mathcal{N} (I,\sigma^2I)
\end{equation}
$X_{emb}$ is the word embedding sequence of the input sentence, $\odot$ is element-wise multiplication.

\subsection{Bernoulli Noise Augmentation}

~\newcite{JMLR:v15:srivastava14a} presented
\emph{Dropout} to tackle the overfitting issue. 
During training, there will be a 
probability of $1 - p$ to replace the value of each word embedding unit with zero value:
\begin{equation}
\label{eq:bernoulli_mul}
X_{emb} \leftarrow  (1/p) X_{emb} \odot e, e \sim \mathcal{B} (n,p)
\end{equation}
where $n$ is the word embedding size.
And there is no dropout during the testing process.
The experimental choice of $p$ could be $0.7 \sim 0.9$ for input signals~\cite{JMLR:v15:srivastava14a}
\footnote{$0.5 \sim 0.75$ for deeper hidden units}.

Dropout is helpful because it can prevent 
features to be co-adapted from each other~\cite{hinton2012improving}
and can also be regarded as a model ensemble method~\cite{JMLR:v15:srivastava14a}. 

\subsection{Adversarial Training}

There is another algorithm called adversarial training~\cite{42503}. 
In the adversarial training, the perturbation is added in the direction of 
maximally increasing the loss function. We apply the approximate version~\cite{goodfellow2014explaining} of adversarial training as follows:
First we replace $X_{emb}$ with $X_{emb} \odot e$, where $e$ is all-one matrix.
Then, we apply adversarial training as:
\begin{equation}
\label{eq:adversarial}
 e \leftarrow  e + \sigma g/\|g\|, g= \nabla _{e} L(X;\theta)
\end{equation}
$L()$ is the loss function, $\theta$ is the model's parameter,
$\nabla _{e}$ is the gradient with respect to $e$,
$\sigma$ is a small hyper-parameter controlling the step size.
Here the new $e$ leads to a small change over $X_{emb}$
which could make the input much more difficult for the model to classify. 
Adversarial training has shown to be effective on the relation 
extraction task~\cite{wu2017adversarial}.

\subsection{Variants with Constraints}

Both Gaussian noise and Bernoulli noise regard noises from different 
units independently. They 
do not consider the factor of output, neither. 
In this section, we introduce some constraints over Gaussian and Bernoulli distribution.

\noindent \textbf{Word Dropout}

Since word should fit in its context, it is often the case that human
can still understand the main meaning of the text  
without reading the whole sentence. So, it is natural to randomly dropout
some words from a text sequence for model training. 

In general, for each word in a sequence, we could dropout the word by
a Bernoulli distribution:

\begin{equation}
\label{eq:word_dropout}
X \leftarrow  X \odot \vec e, \vec e \sim \mathcal{B} (n,p)
\end{equation}
where $X$ is the original word index sequence, 
$n$ is the number of word tokens in the sequence,
and here $\vec e$ is a vector representing dropout states of each
word in a sequence.
In our vocabulary, the word with index zero represents UNK.
Thus, we will replace the word embedding of the dropout word with 
the word embedding of UNK. 
Since the length of the sequence does not change, in this case
there is no need to scale up other word embeddings with $1/p$.

\noindent \textbf{Semantic Dropout}

Different dimensions of word embedding may contain different semantic meanings. 
Under this assumption, we also want the model to recognize each semantic features instead of 
remembering the co-adaptation between them. To emphasize this perspective, we  implemented
dropout on each dimension of word embeddings. Specifically, when one dimension is dropped, 
it means this dimension is dropped by all the words in the sequence.
To be clear, here we need to scale up word embeddings by $1/p$ during training.

\noindent \textbf{Adversarial Noise}

For each gradient descent step, given a certain value of parameters, adversarial training actually provide a
constant change over word embeddings instead of random noises. 
In this case, we would like to combine both Gaussian and Bernoulli noise with
adversarial training.

For \emph{Gaussian adversarial noise}, we first sample $e$ from a Gaussian distribution $\mathcal{N} (I,\sigma^2I)$ instead of 
all-one matrix,
then we apply adversarial training following Formula \ref{eq:adversarial}.

For \emph{Bernoulli adversarial noise}, we first sample $e$ from a Bernoulli distribution $\mathcal{B} (n,p)$.
Different from Gaussian adversarial noise, here we should keep $e_{ij} \in \{0, 1\}$ after adversarial step. 
In this paper, we apply the adversarial dropout algorithm from \newcite{park2017adversarial}. 
Firstly, all the units inside $g$ are sorted (ascending) according to their absolute value. 
Then, in the loop, if $e_{ij} = 1$ and $g_{ij} < 0$, we replace $e_{ij}$ with 0; 
If $e_{ij} = 0$ and $g_{ij} > 0$, we replace $e_{ij}$ with 1.
The loop stops when the number of changes exceeds $ln(1-p)$, where $l$ is the sentence length,
$n$ is the word vector size, $1-p$ is the dropout rate.
And we also need to scale up word embeddings by $1/p$.

\begin{table}[!htb]
\small
\begin{center}
\begin{tabular}{lrrr}
\hline
                     &   Description   & Data size & \# class  \\
\hline
MR      & Movie review~  & 10,662 & 2  \\
        & \cite{pang2005seeing} & & \\
SST2    & The Stanford   & 9,613 & 2 \\
        & Sentiment Treebank  & & \\
        & \cite{socher2013recursive} & & \\
CR      & Customer review & 3,775 & 2 \\
        & \cite{hu2004mining} & &  \\
TREC    & Question type  & 5,952 & 6 \\
        & \cite{li2002learning} & & \\
RE      & SemEval2010 Task8  & 10,717 & 19 \\
        & relation classification & &  \\
        & ~\cite{hendrickx2009semeval} & & \\
TrecQA  & Answer selection & 53,417 & 2 \\
(clean)  & \cite{wang2007jeopardy}  & & \\
\hline
\end{tabular}
\end{center}
\caption{\label{tab:statistics} Dataset Statistics.}
\end{table}

\begin{table*}[!htb]
\small
\begin{center}
\begin{tabular}{|l|l|l|l|l|l|l|r|} 
\hline
\multirow{2}{*}{Model} & MR & SST2 & CR & TREC  & RE &  TrecQA  & Overall\\
\cline{2-7}            &Acc. &Acc. & Acc. &Acc.   & F-1& MAP  & \\
\hline \hline
Baseline               & 0.808 &	0.867 & 0.844 &	0.926 &  0.827 & 0.776 $\pm$ 0.020   & \\
\hline
+ Gaussian Noise      & 0.811 $\uparrow$ &	0.870 $\uparrow$ & 0.845 &	0.925 & 0.827 & 0.775  $\pm$ 0.030 & $-0.1\% \sim 0.3\%$\\
+ Bernoulli Noise     & 0.807 &	0.868 &	0.846 & 0.925 & 0.827 & 0.776 $\pm$ 0.024  & $-0.1\% \sim 0.2\%$\\
+ Adversarial Training * & 0.811 $\uparrow$ & 0.868  & 0.848 $\uparrow$ & 0.928   & 0.828 & 0.780 $\pm$ 0.005 $\uparrow$  & $0.1\% \sim 0.4\%$ \\
\hline
+ Bernoulli Word Noise  & 0.808 &	0.872 $\uparrow$ & 0.843 &	0.926 & 0.826 & 0.774 $\pm$ 0.030  &  $-0.2\% \sim 0.5\%$\\
+ Bernoulli Semantic Noise * & 0.811 $\uparrow$ &	0.872 $\uparrow$ & 0.844 &	0.926 & 0.830  $\uparrow$ & 0.776 $\pm$ 0.033 & $0.0\% \sim 0.5\%$\\
+ Gaussian Adv. Noise * & 0.808 & \textbf{0.873} $\uparrow$ & 0.842  &	0.925 & 0.827 & \textbf{0.786} $\pm$ 0.022 $\uparrow$    &  $-0.2\% \sim 1.0\%$\\
+ Bernoulli Adv. Noise   &  0.808  &  \textbf{0.875} $\uparrow$ & 0.845 & 0.929 $\uparrow$ & 0.826 &   0.737 $\pm$ 0.006 $\downarrow$ &  $-3.6\% \sim 0.8\%$  \\
\hline
\end{tabular}
\end{center}
\caption{\label{tab:results} The performances of different augmentation strategies on
Sentiment/Topic/Relation/Answer classification tasks. * indicates that the method shows 
consistently better or equal performances compared with the original model. $\uparrow$ / $\downarrow$
appears when the difference is equal or larger than 0.3\%. Adv. stands for Adversarial.}
\end{table*}

\section{Experimental Setup}

In this paper, we apply aforementioned methods on several tasks where the training data is limited, 
including sentiment, topic, relation and relevant answer classification. 
Some data statistics are listed in Table~\ref{tab:statistics}.
For the answer selection task, we can
regard it as a binary classification problem where one means the 
question and answer pair are relevant and zero means irrelevant.

For the first five tasks, we chose multi-channel CNN~\cite{kim2014convolutional} with softmax classifier as the 
top model which absorbs the input which is a sequence of word embeddings and provides the output class.
In addition, we added position features~\cite{zeng2014relation} for RE task. 
We used pre-trained word2vec~\cite{mikolov2013distributed} 300 dimension word embeddings for initialization and updated the word embedding during the training process.

For the answer selection task, we employed the multi-perspective CNN 
architecture~\cite{he2015multi} as the top model. 
And we used the GloVe~\cite{pennington2014glove} 300 dimension word embedding
without update following their setup.

We tuned the hyper-parameter on the development set (use cross-validation if there is no development set), 
picked up the model with
the best average performance and evaluated the final performance on the test set.
In practice, we employ $p \in \{0.7, 0.8, 0.9, 0.95\}$, $\sigma \in \{0.001, 0.01, 0.1\}$
in our experiments. In order to observe significant numbers, for each hyper-parameter, 
we trained the model five times and average the performances (10 times for TrecQA).~\footnote{Our code can be
downloaded at \url{https://github.com/zhangdongxu/word-embedding-perturbation}}

\section{Result Analysis}
Table~\ref{tab:results} shows the performances of different perturbation
strategies. Since we observe that the variance of performances on TrecQA dataset is quite huge,
fluctuation ranges are also presented.

From the result, we can see that vanilla adversarial training method is the most safe choice
among different perturbation strategies.
It shows consistent improvements ($0.1\% \sim 0.4\%$) over all the datasets.
Gaussian adversarial noise shows significant improvements on SST and TrecQA datasets.
And Bernoulli-semantic noise is also promising while the performance is not stable on TrecQA dataset.
Another observation is that there are more improvements on some datasets than other ones. But
in most cases, slight perturbation did not hurt the performance.

\subsection{Constraint Effect}

It is interesting to see the influence of different constraints over these
noises.
Adversarial training with Gaussian distribution is better than origin Gaussian noise on these tasks. 
And Bernoulli semantic noise also seems to be superior to vanilla
dropout (Bernoulli noise). These phenomena indicate that certain constraints over
the noise with respect to the application scenario  
may improve models' robustness.

\subsection{Continuous Noise vs Discrete Noise}

Results show that continuous noises such as Gaussian noise and
Gaussian adversarial noise perform equally or slightly better than  
discrete noises such as Bernoulli noise and adversarial dropout. 
This may be because these tasks are
sensitive to certain words and the entire dropout might be too aggressive and could block the model from
learning these features.
From another perspective, 
\newcite{JMLR:v15:srivastava14a} also observed the consistent results. 
They argued that the entropy of a continuous noise could be much higher than a discrete noise, which can be helpful for model learning.

\subsection{Data Size}

\begin{figure}[h]
\includegraphics[width=0.5\textwidth]{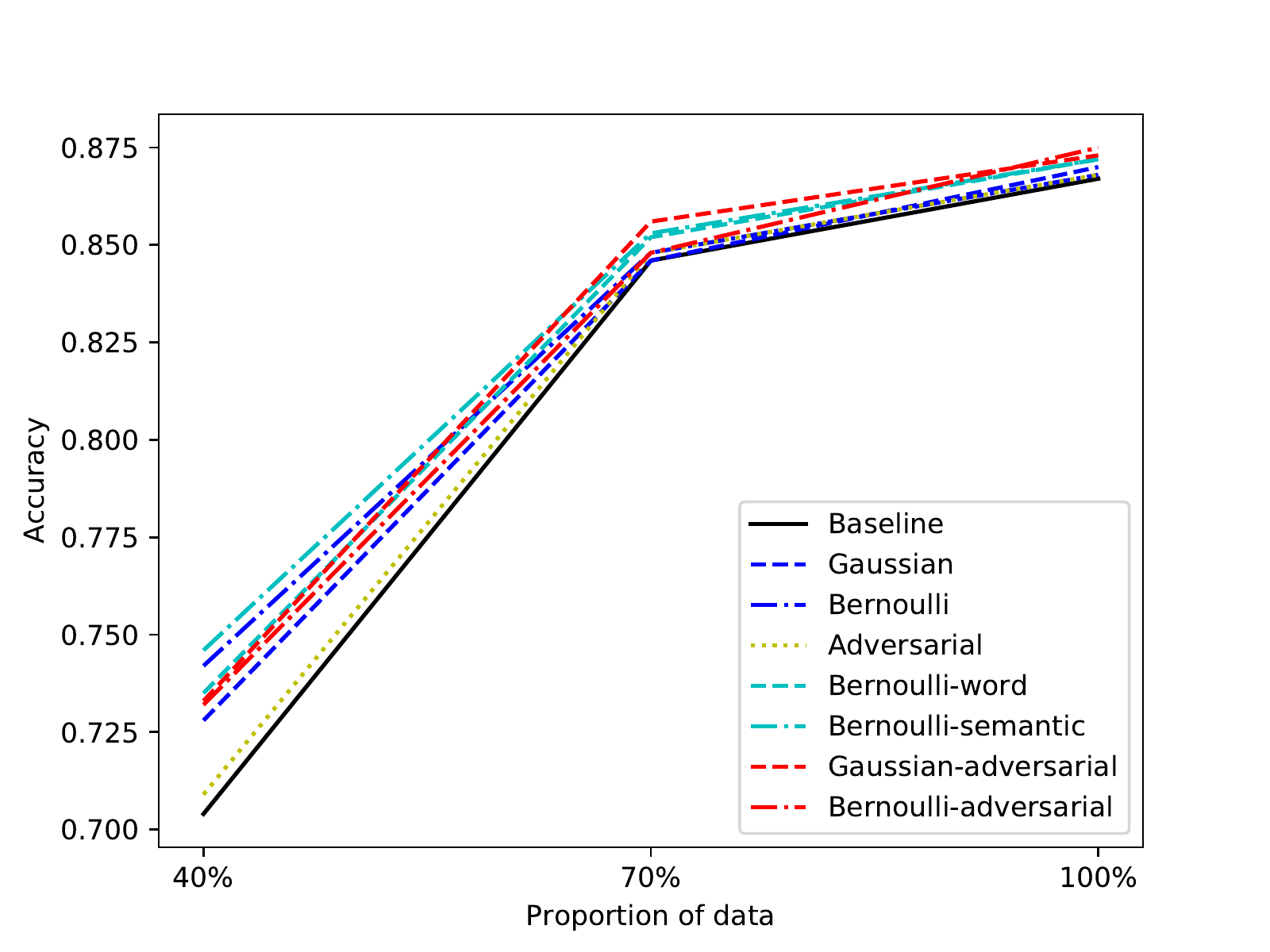}
\caption{Performances on SST2 dataset with different proportions of training data.}
\label{fig:size}
\centering
\end{figure}

To better evaluate the strength of the word embedding perturbation,
we compare performances using different proportions of training
data on the SST2 sentiment analysis task. The testdata results are shown in Figure~\ref{fig:size}.
The figure shows that, comparing with the baseline model, the overall improvement of 
these augmentation methods becomes
larger as the size of training data goes down.


\section{Conclusion}

In this paper,  we compare different word embedding perturbation techniques on sentence classification tasks to tackle the overfitting
issue. 
Experimental results indicate that adding noise on the word embedding layer
can in general improve the model's performance.
And certain types of noise could constantly perform better from
the empirical results.

For the future work, it is interesting to employ task related constraints such 
as word sentiment polarity, or the distances between current word and two entities.
It is also promising but more challenging to accomplish the second strategy mentioned in the introduction. For example we can leverage knowledge bases and language model features 
for multinomial token replacement.

\bibliography{acl2018}
\bibliographystyle{acl_natbib}

\end{document}